\documentclass[10pt,twocolumn,letterpaper]{article}

\usepackage{iccv}
\usepackage{times}
\usepackage{epsfig}
\usepackage{graphicx}
\usepackage{subfig}
\usepackage{floatrow}
\usepackage{amsmath}
\usepackage{amssymb}
\usepackage{multirow}
\usepackage{booktabs}
\usepackage{makecell}
\usepackage{adjustbox}
\usepackage{capt-of,etoolbox}

\graphicspath{ {./image/} }

\usepackage[pagebackref=true,breaklinks=true,letterpaper=true,colorlinks,bookmarks=false]{hyperref}

\iccvfinalcopy 

\def\iccvPaperID{3330} 

\ificcvfinal\fi

\begin{document}
\title{Disentangled Lifespan Face Synthesis}

\makeatletter
\newcommand{\printfnsymbol}[1]{%
  \textsuperscript{\@fnsymbol{#1}}%
}
\apptocmd\@maketitle{{\myfigure{}\par}}{}{}
\makeatother

\author{Sen He\textsuperscript{1,2}\thanks{Equal contribution}, Wentong Liao\textsuperscript{3}\printfnsymbol{1}, Michael Ying Yang\textsuperscript{4}, Yi-Zhe Song\textsuperscript{1,2},  Bodo Rosenhahn\textsuperscript{3}, Tao Xiang\textsuperscript{1,2}\\
\textsuperscript{1}CVSSP, University of Surrey,
\textsuperscript{2}iFlyTek-Surrey Joint Research Centre on Artificial Intelligence,\\
\textsuperscript{3}TNT, Leibniz University Hannover,
\textsuperscript{4}SUG, University of Twente\\

}


\newcommand\myfigure{%
\begin{center}
    \includegraphics[width=0.7\textwidth]{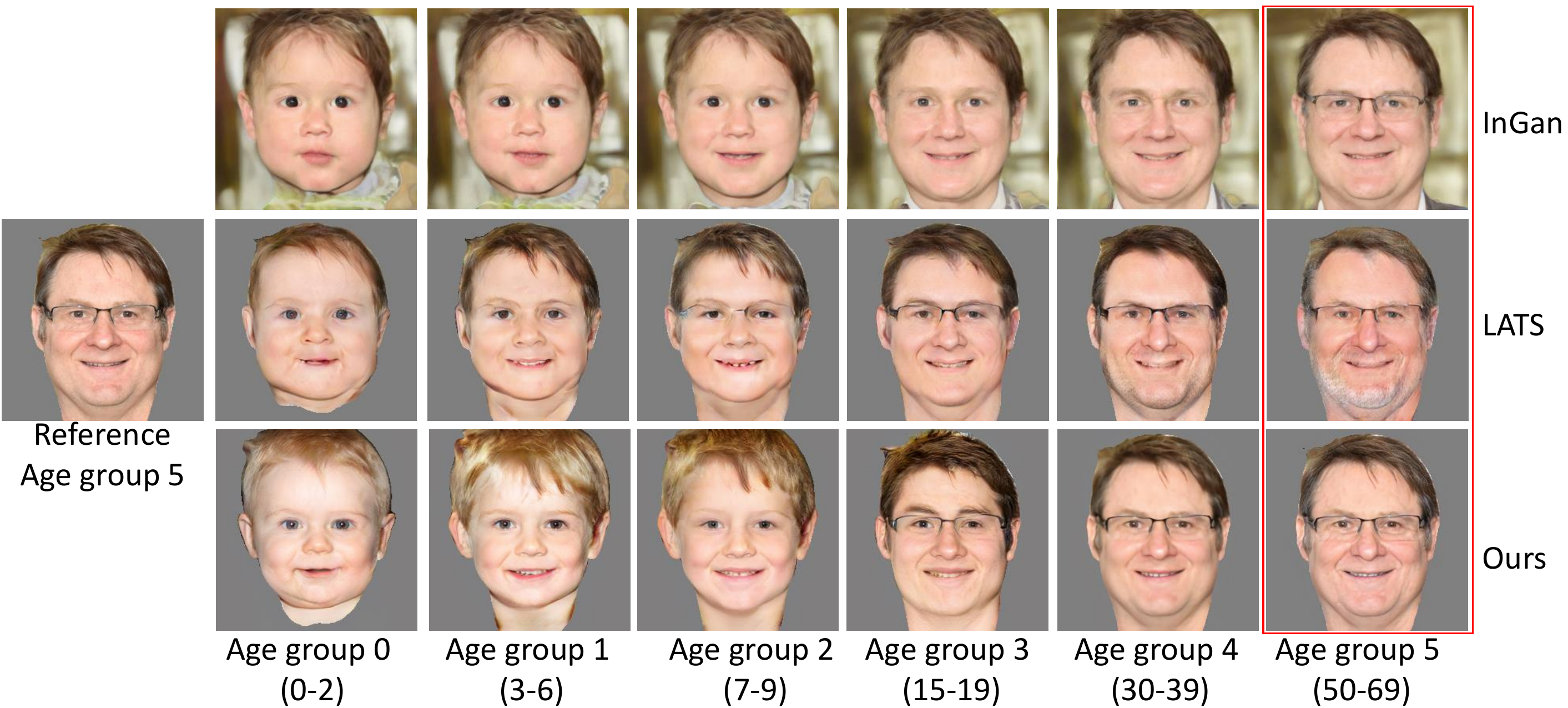}
\captionof{figure}{Examples of generated face images using our lifespan face synthesis model and two state-of-the-art alternatives. InGAN \cite{zhu2020indomain} generates valid texture transformation but fails in shape transformation and identity preservation. LATS \cite{or2020lifespan} improves on shape transformation and identity preservation, but is poor in reconfiguration (the re-generated image in the same age group 5 as the reference looks very different). In contrast, our model overcomes all these limitations, yielding the most plausible effects of aging whilst being identity-preserving. }
\label{fig:fig1}
\end{center}
}

\maketitle
\ificcvfinal\thispagestyle{empty}\fi

\begin{abstract}
A lifespan face synthesis (LFS) model aims to generate a set of photo-realistic face images of a person's whole life, given only one snapshot as reference. The generated face image given a target age code is expected to be age-sensitive reflected by bio-plausible transformations of shape and texture, while being identity preserving. This is extremely challenging because the shape and texture characteristics of a face  undergo separate and highly nonlinear transformations w.r.t.~age. Most recent LFS models are based on generative adversarial networks (GANs) whereby age code conditional transformations are applied to a latent face representation. They benefit greatly from the recent advancements of GANs. However, without explicitly disentangling their latent representations into the texture, shape and identity factors, they are fundamentally limited in modeling the nonlinear age-related transformation on texture and shape whilst preserving identity. In this work, a novel LFS model is proposed to disentangle the key face characteristics including shape, texture and identity so that the unique shape and texture age transformations can be modeled effectively. This is achieved by extracting  shape, texture and identity features separately from an encoder. Critically, two transformation modules, one conditional convolution based and the other channel attention based,  are designed for modeling the  nonlinear shape and texture feature transformations respectively. This is to accommodate their rather distinct aging processes and ensure that our synthesized images are both age-sensitive and identity preserving. Extensive experiments show that our LFS model is clearly superior to the state-of-the-art alternatives. Codes and demo are available on our project website: \url{https://senhe.github.io/projects/iccv_2021_lifespan_face}.
\end{abstract}
\vspace{-0.4cm}

\section{Introduction}
What would a young adult look like as an infant and what will she/he resemble in 20 or even 40 years time? Lifespan face synthesis or face aging and rejuvenation aims to answer these questions by synthesizing the face of a person's whole life given only a snapshot. It is an intriguing problem but also has many applications, \eg cross-age face recognition \cite{park2010age} and finding lost children \cite{wang2016recurrent}. It has therefore attracted a great deal of attention recently \cite{yang2018learning,wang2018face,he2019s2gan,shen2020interpreting,zhu2020indomain, or2020lifespan}.

 Lifespan face synthesis (LFS) is a challenging face attribute editing problem. Compared to other face attribute editing works \cite{he2019attgan, liu2019stgan,yin2019instance} where many attributes such as glasses, hair style and smiling are manipulated with a single model, LFS focuses on one attribute only, namely age. However, age editing is arguably the hardest task out of all the attributes. It is thus typically studied on its own. This is because aging is an extremely complex face transformation process. In particular, over the lifespan of a person, the face experiences changes in both shape and texture \cite{lgfarkas}. Further, such changes are nonlinear over time and rather different for shape and texture: a face's appearance changes are first dominated by shape deformation from baby to young adult because of the growth of bones of skull; such changes then primarily take the form of texture transformation when an adult grows older, \eg colors of beard and hair, wrinkles. 

Therefore, an ideal LFS model must meet three requirements \cite{he2019s2gan,shen2020interpreting,zhu2020indomain, or2020lifespan}: (1) {\bf Age-sensitive reflected by bio-plausible shape and texture transformations}:  Given a reference face image and a random target age, the generated face image should have valid shape deformation as well as texture transformation compared to the reference face image. In particular, the highly nonlinear transformations mentioned above need to be respected.  (2) {\bf Identity-preserving}: no matter how large the age gap is between the target and reference, the generated image must depict the same person. (3) {\bf Reconfigurable}: When the target age is the same as the age of the reference face image, the generated face image should be as similar to the reference image as possible. 



However, despite the best efforts of researchers in the past two decades, none of the existing LFS models can meet all three requirements. Before the deep learning era, LFS models are either `prototype' based  \cite{tiddeman2001prototyping, tazoe2012facial, kemelmacher2014illumination} modeling mean age appearance for different age groups, or `physical' based \cite{suo2009compositional, tazoe2012facial} with explicitly modeling of the underlying biological aging mechanism. The former omits personalized information. While the later one  requires images of same persons over the whole lifespan which is infeasible to scale.  
More recent methods \cite{wang2016recurrent, nhan2017temporal,antipov2017face, zhang2017age, yang2018learning,wang2018face,he2019s2gan,shen2020interpreting,zhu2020indomain, or2020lifespan} benefit from the advancements in deep generative adversarial networks (GANs) \cite{goodfellow2014generative, arjovsky2017wasserstein}. Using these methods, a latent face representation encompassing information of shape, texture and identity is transformed conditional on the target age before being fed into an image generator. Thanks to the recent breakthroughs in GANs such as Style-GAN \cite{karras2019style, karras2020analyzing}, these models can now generate incredibly high quality face images. But as shown in  Fig.~\ref{fig:fig1}, they still fail in one or more of the three requirements.

This is because none of these models can effectively disentangle a face representation into shape, texture and identity related parts. Such a disentanglement is crucial for LFS because without the disentanglement, it is impossible to apply different age-conditional manipulations to these different representations to model the aforementioned nonlinear transformations in shape and texture appearance, whilst being identity-preserving. 
As a result, it is difficult to avoid unwanted editing. For example, identity can be changed  as illustrated in the first row in Fig.~\ref{fig:fig1}. Furthermore, some incompatible transformation may occur, yielding unrealistic effects in the generated images (glasses start to appear in age group 2 of the middle row example).

In this work, for the first time we propose a LFS model that explicitly disentangles a learned latent face representation into  shape, texture and identity. Our model is a conditional GAN with an encoder-decoder architecture. First,  features of different layers of a shared CNN encoder are extracted and subject to different feature extraction modules. Second, to model the distinct nonlinear transformations on shape and texture with respect to age, two novel feature transformation modules are developed for shape and texture. They are based on conditional convolution and channel attention respectively to reflect the intrinsically different aging effects on shape and texture.  Last but not least,  to facilitate the disentanglement of  shape and texture, a regularization loss is introduced on shape based on the intuition that shape changes are small when an adult is growing older \cite{setiawati2019bone}. As shown in Fig.~\ref{fig:fig1}, our disentanglement LFS model can effectively overcome the limitations of the state-of-the-art competitors and meet all three requirements simultaneously.  

\textbf{The contributions} of this paper are as follows: (1) for the first time, we explicitly model the face's shape, texture and identity characteristics in  an end-to-end trained lifespan face synthesis (LFS) model. (2) To model the separate nonlinear aging processes on shape and texture, we propose separate shape and texture transformation modules based on  conditional convolution and channel attention respectively, as well as  a shape regularization loss  to facilitate  the disentanglement. (3) Extensive experiments are carried out to demonstrate that our model is much superior to the state-of-the-art alternatives.

\section{Related Works}
\paragraph{Generative adversarial networks}
Generative adversarial networks (GANs) \cite{goodfellow2014generative} are used by most recent image generation and manipulation methods. The developments of GANs can be mainly divided into two groups since the vanilla GANs \cite{goodfellow2014generative}. One group tries to better measure the distribution divergence between the generated images and the original images \cite{arjovsky2017wasserstein}. 
The other group focuses on the architecture design, which has evolved from the original fully connected networks to multi-scale convolutional architectures \cite{karras2017progressive}. The most recent architectures is the style-GAN architecture \cite{karras2019style,karras2020analyzing}, where a random noise is first projected into a latent space and then used for convolution modulation. Style-GAN architecture has also been adopted in the recent state-of-the-art lifespan face synthesis models \cite{shen2020interpreting,or2020lifespan,zhu2020indomain} as well as the proposed model in this work. 
\vspace{-0.4cm}
\paragraph{Face manipulation}
Face manipulation aims to edit a reference face image by changing some attributes, \eg, age, smile and pose. The manipulated image is expected to contain intended attribute changes whilst preserving other attributes and identity.  Recently, face manipulation has been studied intensively \cite{he2019attgan, liu2019stgan,yin2019instance,  shen2020interpreting, he2020pa, kwak2020cafe, zhu2020indomain}. AttGAN \cite{he2019attgan} uses an attribute classification constraint to regularize the manipulated image. STGAN \cite{liu2019stgan} selectively transfers the required attribute while keeping other factors unchanged. \cite{shen2020interpreting} learns the direction of each attribute in the latent space of style-GAN \cite{karras2019style, karras2020analyzing}, and then manipulates the exact latent code \cite{zhu2020indomain} of the reference image accordingly. 
Note that some generic face manipulation models do support age editing. However, they only manipulate a face image to be either younger or older (i.e., binary manipulation), an easier task than LFS. 
\vspace{-0.4cm}
\paragraph{Lifespan face synthesis}
Lifespan face synthesis (LFS) is the most challenging face manipulation task. 
Classical `prototype' based methods \cite{tiddeman2001prototyping, tazoe2012facial, kemelmacher2014illumination}  divide the continuous ages into several discrete age clusters, and then compute the mean face in each cluster for reference. 
In contrast, `physical' based methods model the change of each aging factor in a parametric manner. \cite{lanitis2002toward} explored different parametric models (linear, quadratic and cubic) for the aging function. \cite{suo2012concatenational} uses a concatenational graph to model the aging process. Both groups of methods are based on manually designed rules, which is impossible to approximate the complex and nonlinear aging process. Further, they usually require images of the same person at different ages, which are very difficult to collect.

More Recent methods  use conditional GANs for image generation. Yang \etal \cite{yang2018learning} propose a pyramid discriminator to penalize different factors in the aging progress. IPGAN \cite{wang2018face} uses AlexNet \cite{krizhevsky2012imagenet} pre-trained on ImageNet \cite{russakovsky2015imagenet} to enhance the identity preservation in the aging process. S2GAN \cite{he2019s2gan} learns different transformation basis for different age groups. LATS \cite{or2020lifespan} employs style-GAN architecture, in which the input is the encoded reference image and the style code is the embedded age representation. All these methods apply age-conditional transformation on an entangled latent representation of the reference image. They are thus intrinsically limited in modeling the distinct nonlinear transformations of shape and texture over ages. This limitation motivates the proposed disentanglement LFS approach. 
\vspace{-0.8cm}
\paragraph{Face disentanglement}
There are many existing efforts on disentangling the face into different latent factors, \eg, identity, pose, shape and texture.
Peng \etal \cite{peng2017reconstruction} propose to disentangle face into identity and pose by reconstruction. Shen \etal \cite{shen2020interpreting} propose to disentangle the learned latent space into different attributes by supervised projection. Nitzan \etal \cite{nitzan2020face} disentangle the identity information via latent space mapping. In this work, we propose to disentangle face into two aging-related factors, \ie, shape and texture, as well as age-insensitive factor, namely identity. The key novelties over existing face disentanglement works are the two separate transformation modules designed to capture the distinct aging effects on shape and texture, and a shape regularization loss. These are crucial for effective age-sensitive disentanglement. 
\vspace{-0.4cm}
\paragraph{Dataset for lifespan face synthesis} is  hard to collect because ideally it should contain face images of the same person from baby to pensioner age. Most existing datasets miss face images of the 0 to 10 years age range. For example, the popular MORPH dataset \cite{ricanek2006morph} only has an age range of  16 to 77. The only dataset covering the whole range is the recently re-annotated FFHQ dataset \cite{or2020lifespan}, which contains ages ranging  from 0 to 70 years old. It is thus used in this work. Overall existing LFS datasets are relatively small and insufficient for fine-grained age synthesis.

\begin{figure*}[ht!]
    \centering
    \includegraphics[width=0.99\textwidth]{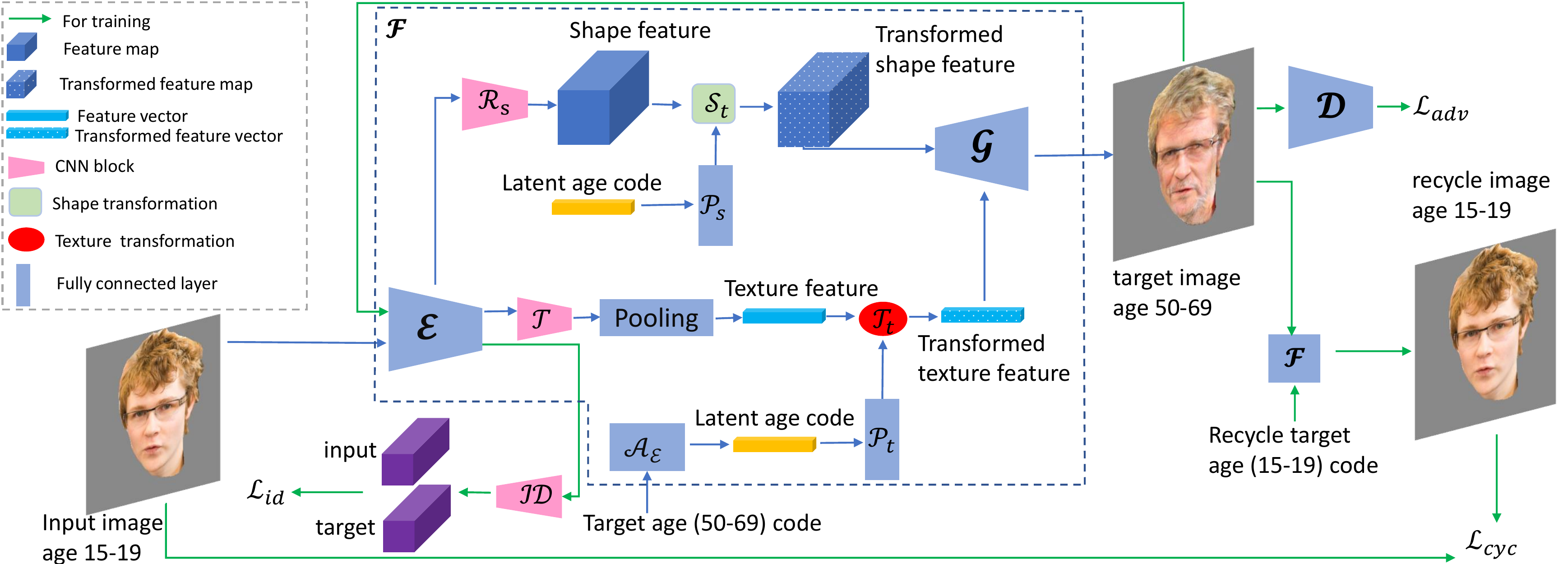}
    \caption{A schematic of our model. A latent representation of  a reference image is disentangled into shape and texture relevant features, which are then transformed through separate transformation modules, conditioned on the target age. The transformed shape and texture feature are then fed into a style-GAN generator for the target image generation.}
    \label{fig:pipeline}
        \vspace{-2mm}
\end{figure*}

\section{Method}
\subsection{Problem definition}

Due to the dataset limitation, we follow \cite{or2020lifespan} and divide the age range into 6 discrete groups (0 to 5). Each group has an age code $z_{i} \in \mathbb{R}^{\lvert6N \rvert}$, which is computed as:
\begin{equation}
    z_{i} = \mathbf{1}_{i} + \mathbf{n},
\end{equation}
where $\mathbf{1}_{i}$ contains all ones on elements through $iN$ to $(i+1)N$ and zeros elsewhere, $\mathbf{n}\in\mathbb{R}^{\lvert6N\rvert}$ is a Gaussian noise.

Given a reference face image $I_r \in \mathbb{R}^{H \times W \times 3}$ from the $r_{th}$ age group and a target age code $z_t$ for the $t_{th}$ age group, a lifespan face synthesis model $\mathcal{F}$ aims to generate a target face image $I_t$, where $I_{t} = \mathcal{F}(I_r, z_t)$. The generated face image should have the same identity as the reference image but exhibit age-sensitive  texture and shape changes according to the target age.
Discrete lifespan face synthesis is done by traversing the age code of all age groups. To synthesize faces in a more fine-grained manner, the corresponding age code can be obtained by linear interpolation between the age codes of the neighboring two age groups.

As illustrated in Fig.~\ref{fig:pipeline}, our model  consists of five parts, an encoder ($\mathcal{E}$), a shape transformation module ($\mathcal{S}_{t}$), a texture transformation module ($\mathcal{T}_{t}$), an age embedding module ($\mathcal{A_E}$), a generator ($\mathcal{G}$) and a discriminator ($\mathcal{D}$). Each of them will be detailed in the following sections.

\subsection{Feature extraction}
\label{susec:feature}

Previous methods \cite{wang2018face,zhu2020indomain,or2020lifespan} extract entangled representation of face image and transform it according to the target age. Without factorizing the latent representation into shape, texture and identity relevant factors, it is impossible to model the transformations on shape and texture separately whilst preserving identity. Latent representation disentanglement is thus the key to effect LFS meeting all three requirements. To that end,  we use our encoder ($\mathcal{E}$) to extract 3 distinct sets of  features, \ie, shape ($f_{s}$), texture ($f_{t}$)  and identity ($f_{id}$). Inspired by the neural style transfer works \cite{gatys2015neural,johnson2016perceptual} which suggest that structure information can be extracted from the middle layers of a CNN and texture information from the deeper layers, we also propose to extract these three features from different layers of the encoder CNN ($\mathcal{E}$).

Concretely, the shape features ($f_{s} \in \mathbb{R} ^{ \frac{H}{4} \times \frac{W}{4} \times C}$) are extracted from the middle part of our encoder ($\mathcal{E}_{m}$):
\begin{equation}
    f_{s} = \mathcal{R}_{s}(\mathcal{E}_{m}(I_{r})),
\end{equation}
where $\mathcal{R}_s$ is a residual block \cite{he2016deep} to extract shape information for the raw features from the middle part of CNN. Both texture and identity features are extracted from the last layer of our encoder ($\mathcal{E}_{d}$). Specifically, the texture features ($f_{t} \in \mathbb{R}^{\lvert C\rvert}$) are computed as:
\begin{equation}
    f_{t} = \mathcal{T}(\mathcal{E}_{d}(I_{r})),
\end{equation}
where $T$ is a convolutional projection module that extracts the texture information and pools it into a vector. In the meanwhile, the identity features ($f_{id} \in \mathbb{R} ^{\frac{H}{8} \times \frac{W}{8} \times 2C}$) are extracted by another convolutional projection module ($\mathcal{ID}$) with one downsampling layer:
\begin{equation}
    f_{id} = \mathcal{ID}(\mathcal{E}_{d}(I_{r})).
\end{equation}

\subsection{Shape and texture transformation}
After feature extraction, we model the age-conditioned shape and texture transformation in a different way. For the shape transformation, we use conditional convolutions, where the convolution filters are modulated by the target age information:
\begin{equation}
    \begin{aligned}
    f_{s}(z_{t}) & =  \mathcal{S}_{t}(f_{s},z_{t}) \\
    & = \mathbf{conv}(f_{s}, \mathcal{M}(\mathbf{w}_{s}, \mathcal{P}_{s}(\mathcal{A_{E}}(z_{t}))), 
    \end{aligned}
\end{equation}
where $\mathbf{w_{s}}$ is the filter weights, $\mathcal{A_{E}}(z_{t}) \in \mathbb{R}^{\lvert C\rvert}$ encodes the target age code into a latent space, which is used for convolution modulation ($\mathcal{M}$), and $\mathcal{P}_{s}$ is a linear projection layer. With our formulation, the target age information will modulate the filter's weights and thus implicitly change shape information accordingly. Conditional convolution is adopted because shape transformation over age is often global and progressive; age code conditioned convolutional filters are well suited for capturing such transformation. Further, it is also flexible enough for learning at what age group shape changes become minimal. 

For texture transformation, we use the age-conditioned channel attention, defined as: 
\begin{equation}
    f_{t}(z_{t}) = \mathcal{T}_{t}(f_{t}, z_{t}) = f_{t} \circ \mathcal{P}_{t}(\mathcal{A_{E}}({z_{t}})),
\end{equation}
where $\circ$ is element-wise multiplication and $\mathcal{P}_{t}$ is a linear projection layer. Again, this is determined by the nature of the texture changes caused by aging. In particular, as shown in \cite{shen2020interpreting,zhu2020indomain}, different elements in face features $f_{t}$ represent different attributes, \eg, hair color and wrinkles. These attribute are present across ages but with different strengths.   With age-conditioned channel attention, different aging attributes can be easily amplified or suppressed. For example, the attention module will learn that wrinkles need to be suppressed by younger ages while amplified by older ages.

The transformed shape and texture information are then fed into a style-GAN \cite{karras2020analyzing} based generator $\mathcal{G}$ for target image generation:
\begin{equation}
    I_{t} = \mathcal{G}(f_{s}(z_{t}), f_{t}(z_{t})).
\end{equation}

\subsection{Shape regularization}
Inspired by the previous finding \cite{lgfarkas} that the shape of an adult face usually remains unchanged,  we propose a shape regularization to enforce this observation. This regularization would indirectly ensure that  the extracted features $f_{s}$ is indeed shape related and disentangled from the texture feature $f_{t}$ which changes significantly when an adult is getting older. Concretely, for transforming a reference face image $I_{r_{e}}$ in an adult group $r_{e}$ (\eg 40 years old)  into an older age group $t_{e}$ (\eg 60 years old),  the transformed face image $I_{t_{e}}$ should have the same shape information:
\begin{equation}
    \mathcal{L}_{s} = \left\Vert \mathcal{R}_{s}(\mathcal{E}_{m}(I_{r_{e}})) - \mathcal{R}_{s}(\mathcal{E}_{m}(I_{t_{e}})) \right\Vert^2,
    \label{eq:shapereg}
\end{equation}
where $\mathcal{L}_{s}$ measures shape difference and will be minimized. 

\subsection{Objectives}

There are 5 learning objectives in our model's training. 
To ensure the identity preservation, an identity loss $\mathcal{L}_{id}$ is computed using the identity information between the reference image and the generated target image:
\begin{equation}
    \mathcal{L}_{id} = \left\Vert \mathcal{ID}(\mathcal{E}_{d}(I_{r})) - \mathcal{ID}(\mathcal{E}_{d}(I_{t})) \right\Vert^2.
\end{equation}
Meanwhile, a cycle consistency loss is applied to enhance the identity preservation:
\begin{equation}
    \mathcal{L}_{cyc} = \left\Vert I_{r} - \mathcal{F}(I_{t}, z_{r})\right\Vert^2.
\end{equation}
To maintain the model's reconfiguration, a reconstruction loss is used when the target age is the same as the reference age:
\begin{equation}
    \mathcal{L}_{r} = \left\Vert I_{r} - \mathcal{G}(f_{s}(z_{r}), f_{t}(z_{r}))\right\Vert^2.
\end{equation}
Furthermore, a conditional adversarial loss is used to improve the realism of the generated images:
\begin{equation}
    \begin{aligned}
    \mathcal{L}_{adv} = & \  \mathbb{E}_{I_{im}^{r} \sim p_{data}^{r}(I_{im}^{r})}[log(\mathcal{D}(I_{im}^{r} | z)] \\
    & + \mathbb{E}_{I_{im}^{g} \sim p_{data}^{g}(I_{im}^{g})}[1-log(\mathcal{D}(I_{im}^{g} | z)].
    \end{aligned}
\end{equation}
The overall training objectives are summed together:
\begin{equation}
    \mathcal{L} = \lambda_{adv}\mathcal{L}_{adv} + \lambda_{r}\mathcal{L}_{r} + \lambda_{cyc}\mathcal{L}_{cyc} + \lambda_{id}\mathcal{L}_{id} + \lambda_{s}\mathcal{L}_{s},
\end{equation}
where $\lambda_{adv}$, $\lambda_{r}$, $\lambda_{cyc}$, $\lambda_{id}$, and $\lambda_{s}$ denote the hyperparameters for balancing the 5 objectives.

\begin{figure*}[t!]
    \centering
    \includegraphics[width=1\textwidth]{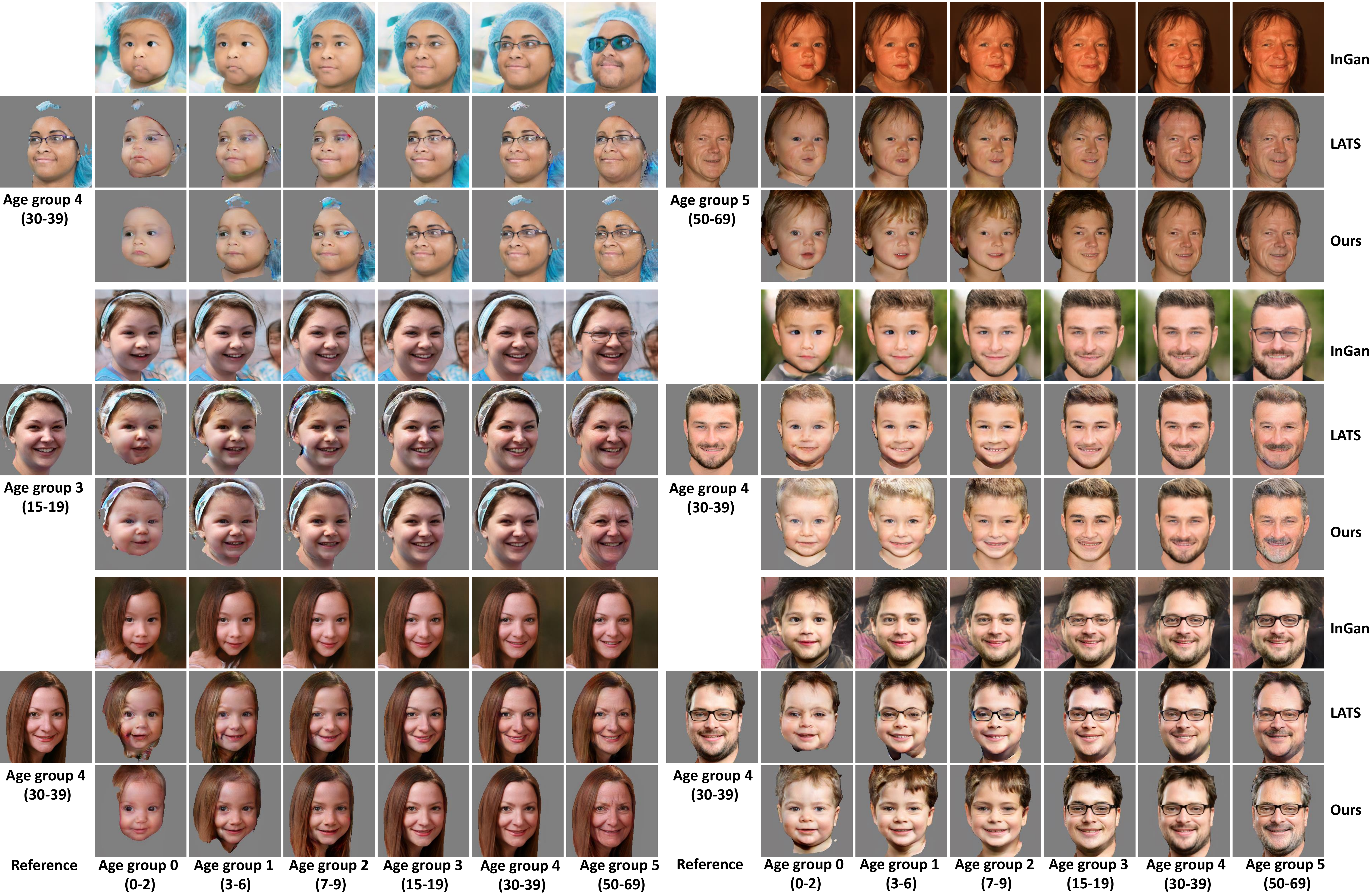}
    \caption{Qualitative results comparing our model against recent state-of-the-art models InGan \cite{zhu2020indomain} and LATS \cite{or2020lifespan}.}
    \vspace{-0.2cm}
    \label{fig:main_quali}
\end{figure*}

\begin{table*}[th!]
    \centering
    \vspace{-2mm}
    \begin{tabular}{c|c|c|c|c|c|c}
    \toprule
         \multirow{2}{*}{Methods}& \multirow{2}{*}{\shortstack{Identity \\ preservation}$\uparrow$} & \multirow{2}{*}{\shortstack{Shape \\ transformation}$\uparrow$} & \multirow{2}{*}{\shortstack{Texture \\ transformation}$\uparrow$} & \multirow{2}{*}{Reconfiguration$\uparrow$} &\multirow{2}{*}{\shortstack{Age \\ error}$\downarrow$} &\multirow{2}{*}{\shortstack{Age \\ accuracy}$\uparrow$}\\ 
         &&&&&\\
         \midrule
         IPGAN \cite{wang2018face}&\textbf{3.92$\pm$0.17}&2.38$\pm$0.42&2.50$\pm$0.12&3.93$\pm$0.01&11.33$\pm$0.89&27.0\%\\
         InGAN \cite{zhu2020indomain}   &2.74$\pm$0.17&2.51$\pm$0.22&2.37$\pm$0.16&3.56$\pm$0.35&8.64$\pm$2.80&39.4\%\\
         LATS \cite{or2020lifespan}     &3.18$\pm${0.13}&2.89$\pm$0.44&3.22$\pm$0.17&3.49$\pm$0.25&5.67$\pm$3.61&60.0\%\\
         Ours                           &3.07$\pm$0.19&\textbf{3.18}$\pm$\textbf{0.35}&\textbf{3.30}$\pm$\textbf{0.21}&\textbf{4.07}$\pm$\textbf{0.27}&\textbf{3.53}$\pm$\textbf{2.81} &\textbf{65.6\%}\\
    \bottomrule
    \end{tabular}
    \caption{User study results for different compared models.}\label{tab:manual_e}
    \vspace{-3mm}
\end{table*}

\section{Experiments}
\vspace{-0.2cm}
\paragraph{Datasets} We train our model on the current largest face age dataset, \ie FFHQ dataset \cite{karras2019style},  with age annotations covering all age groups. 
Due to label noise in the annotation, the original dataset with 70000 images is pruned to 28701 images~\cite{or2020lifespan}. 
The pruned dataset has 14232 training images and 198 testing images for male, and 14066 training images and 205 testing images for female. As per standard, we train male's model and female's model separately. Ages are divided into 6 discrete age groups (0-2, 3-6, 7-9 , 15-19, 30-39, 50-59) in the training dataset. The last two age groups are used to apply the shape regularization in Eq.~(\ref{eq:shapereg}). Following \cite{or2020lifespan}, the non-face regions in the input images are masked out using off-the-shelve face parsing model \cite{chen2017rethinking} trained on CelebAMask-HQ \cite{lee2020maskgan}.
\vspace{-0.5cm}
\paragraph{Implementation Details} Our model is implemented in PyTorch. Each input image is resized to $256 \times 256$, the same size as in \cite{or2020lifespan}. We set the batch size to 2 given the hardware available to us (a single Nvidia RTX 2080-Ti GPU). The length of age code is set to 300 ($N=50$). The latent space dimension $C=256$. The encoder $\mathcal{E}$ has two pooling layers in the first two blocks. And the generator $\mathcal{G}$ has two upsampling layers in the last two blocks. All parameters are trained using Adam optimizer \cite{kingma2014adam}. The initial learning rate is set to 0.001, and decayed by 0.1 at epoch 50 and 100. The whole model is trained with 300 epochs. EMA \cite{yazici2019unusual} is used in the model training. 
\vspace{-0.5cm}
\paragraph{Evaluation Metrics} We evaluate our model both automatically and manually (user study). In automatic evaluation, we use the off-the-shelve VGG-face \cite{parkhi2015deep} to evaluate the model's identity preservation. We also use LPIPS \cite{zhang2018unreasonable} to evaluate model's reconfiguration between the reference image and the re-generated image in the same age group. For manual evaluation, we run perceptual study on Amazon
Mechanical Turk (AMT) to compare the quality of the generated lifespan face images from different models. Given a reference image, the generated images using different models are evaluated from 6 perspectives, namely, \textit{{identity preservation, shape transformation, texture transformation, reconfiguration, age error and age accuracy}}. For identity preservation, the AMT workers were asked to judge how well the generated images preserved the identity in the reference image. For shape and texture transformation, they scored how plausible the transformations are. For reconfiguration, a worker was asked to score how similar the generated image in the same age group is to the reference image. {{For these 4 metrics, the scores are in 5 levels (1-5, higher being better)}}. For age error and age accuracy, each worker was asked  to evaluate whether the generated image belongs to the target age group and estimate the age difference to that group.  Each AMT worker was randomly allocated 30 reference images. 10 workers participated in the evaluation for all 6 metrics.
\vspace{-0.6cm}
\paragraph{Baselines} We compare our model with two state-of-the-art lifespan face synthesis models, \ie, LATS \cite{or2020lifespan} and InGAN \cite{zhu2020indomain}, both using styel-GAN based generators as ours. To synthesize lifespan faces in InGAN, the aging parameter is adjusted according to the reference images' age. We also compare with IPGAN \cite{wang2018face}, which uses a standard convolutional neural network as generator with a focus on identity preservation.


\begin{figure*}[t!]
\centering
\includegraphics[width=0.7\textwidth]{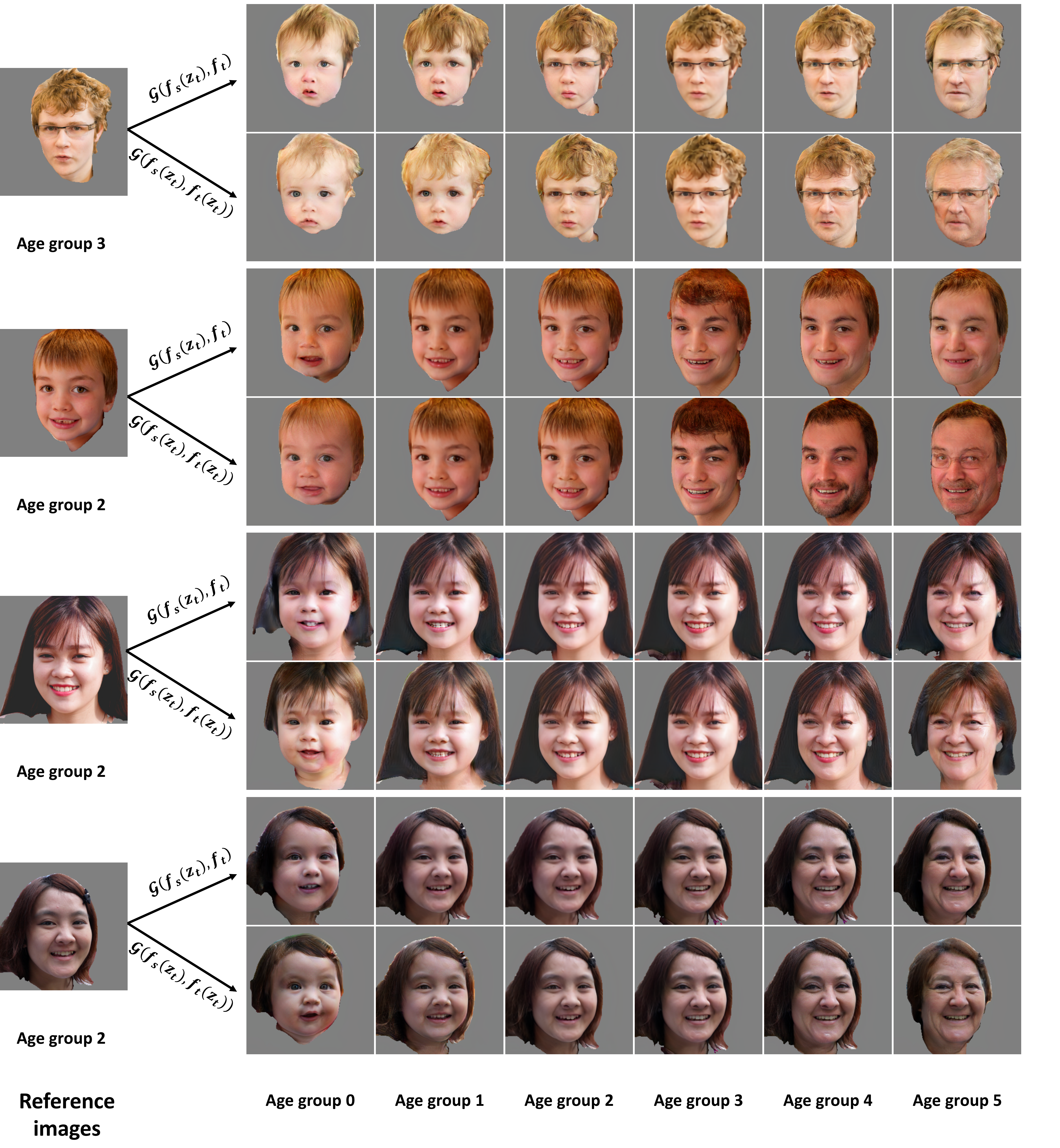}
\caption{Validation of the disentanglement of  shape and texture features in our model. In the first row of each example, the reference's texture information is fixed. Only shape information is transformed to synthesize the face images in different age groups. In the second row of each example, texture transformation is added to contrast its aging effects against shape.}
\label{fig:shape_texture}
\vspace{-0.4cm}
\end{figure*}

\subsection{Main results}
The results of user study and automatic evaluation are shown in in Tables~\ref{tab:manual_e} and~\ref{tab:auto_e}, respectively.
It can be seen that our model achieves significantly better overall performance using all evaluation metrics. It is noted that IPGAN \cite{wang2018face} has very good identity preservation but fails completely at the main task, i.e., generating age-sensitive bio-plausible shape and texture transformations.
As a result, its reconfiguration results are very strong in both tables. However, this is due to the fact that it hardly changes the face regardless what the target age is. LATS \cite{or2020lifespan} is capable of good texture transformation, but is  poor on reconfiguration and yields much lower age accuracy than our model. Note that for LFS, the identity preservation objective is often contradictory to those of  shape and texture transformation. 
Table~\ref{tab:manual_e} shows that our model's identity preservation is marginally lower than LATS, but this is more then compensated by the much more superior performance on the other 5 metrics.  
We can see from the qualitative comparison in Fig.~\ref{fig:main_quali} that our model can synthesize lifespan face with (1) more significant shape deformation among young groups and better texture transformation among older groups, (2) better reconfiguration, and (3) more realistic images (\eg, baby's face and glasses).

\begin{table}[t!]
    \centering
    \small\addtolength{\tabcolsep}{0pt}
    \caption{Automatic evaluation on identity preservation (ID) and reconfiguration (Reconfig).}
    \vspace{-0.5cm}
    \begin{tabular}{c|c|c}
    \toprule
         Methods& ID ($\uparrow$) & Reconfig ($\downarrow$) \\ \midrule
         IPGAN \cite{wang2018face}&\textbf{99.18}\%& \textbf{0.03 $\pm$ 0.01} \\
         InGAN \cite{zhu2020indomain}&92.35\%&0.17$\pm$ 0.09\\
         LATS \cite{or2020lifespan}&96.68\%&0.11 $\pm$ 0.03\\
         Ours&96.47\%& 0.07 $\pm$ 0.02  \\
    \bottomrule
    \end{tabular}
    \label{tab:auto_e}
    \vspace{-0.3cm}
\end{table}
\subsection{Ablation study}
\vspace{-0.1cm}
\paragraph{Disentangled Representations} Our main idea is to learn disentangled representations. Did our model learn it? In this section, we qualitatively check, from two perspectives, whether the learned shape and texture representations have indeed been disentangled. First, we generate lifespan face images by only transforming $f_{s}$ while keeping $f_{t}$ fixed, i.e., $I_{t} = \mathcal{G}(f_{s}(z_{t}), f_{t})$. We then transform both $f_{s}$ and $f_{t}$ to contrast their functionalities, \ie, $I_{t} = \mathcal{G}(f_{s}(z_{t}), f_{t}(z_{t}))$. The result is shown in Fig.~\ref{fig:shape_texture}. It can be seen that the transformation of $f_{s}$ yields significant shape deformation in the generated faces. However, it has little impact on the texture of the generated faces. Once the transformation of $f_{t}$ is added in, we can see then significant texture changes from young to older adults  (age group 4 to 5). Interestingly, the added $f_{t}$ has little impact on the shape of the generated images. These results therefore validate the design  that $f_{s}$ and $f_{t}$ are learning shape and texture information, respectively. More importantly, shape transformation is indeed more significant in younger groups (age group 0 to 3) while the added texture transformation mostly influences the texture in the older groups (age group 4 to 5). In other words,  the nonlinear aging process has been learned in our proposed shape and texture transformation.

\begin{figure}[t!]
\centering
\vspace{-5mm}
\includegraphics[width=0.95\textwidth]{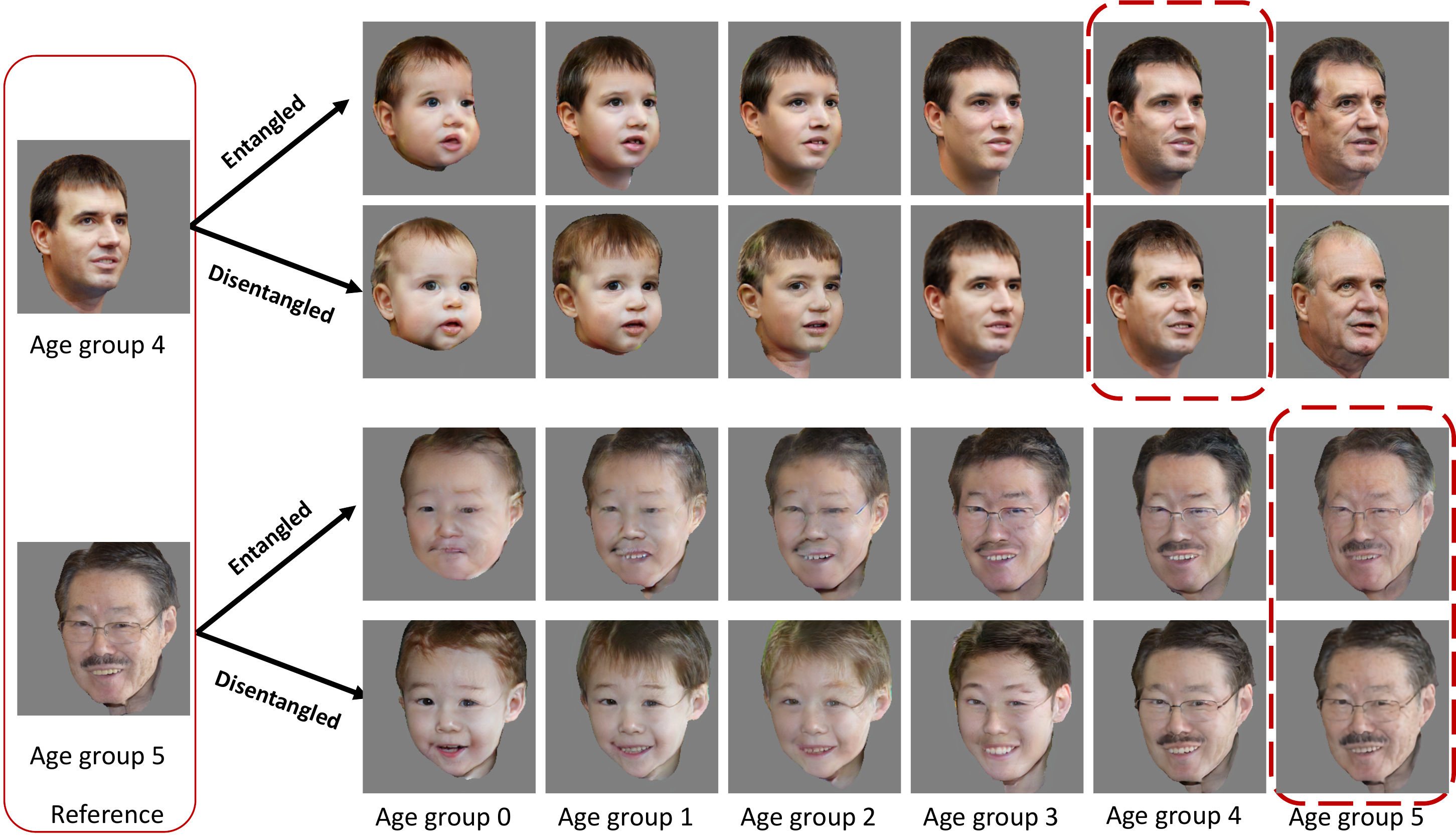}
\caption{Qualitative comparison between entangled and disentangled lifespan face synthesis. In each example, the top row is the entangled model while the bottom row is disentangled. Red dotted boxes indicate that the generated images are within the same age group as the reference image.}
\label{fig:endis}
\vspace{-0.3cm}
\end{figure}

\vspace{-0.4cm}

\begin{figure}[t!]
\centering
\includegraphics[width=0.95\textwidth]{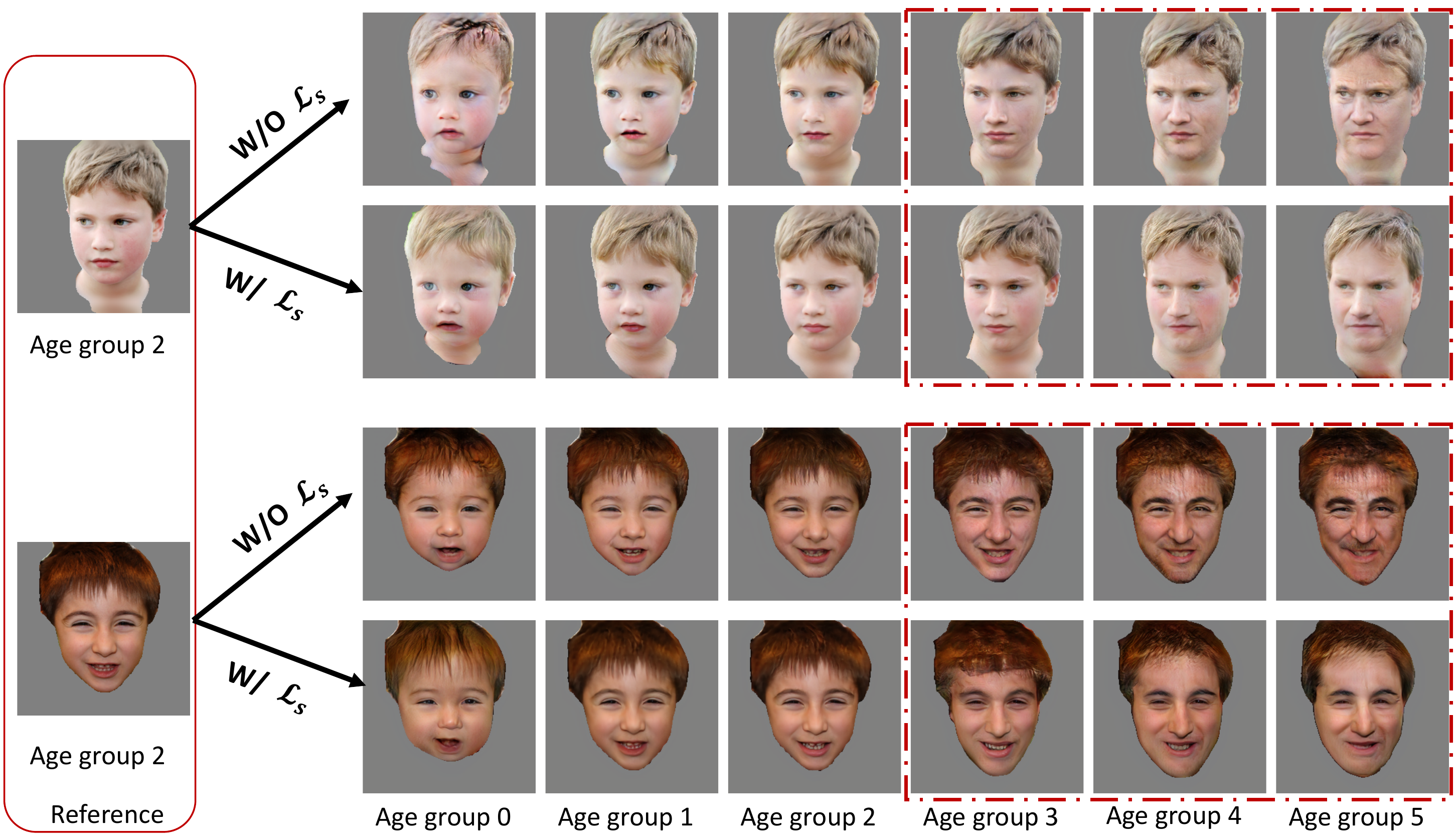}
\caption{Qualitative comparison between model w/o and w/ shape regularization ($\mathcal{L}_{s}$). }
\vspace{-0.4cm}
\label{fig:wswos}
\end{figure}
\paragraph{Disentangled vs.~Entangled Representation}
How important is disentanglement for lifespan face synthesis? To answer this question, we  use the same encoder in our model to extract an entangled feature representation $f_{en}$ for the reference image. We then generate the target image using the same generator, conditioned on the target age, \ie, $I_{t} = \mathcal{G}(f_{en}, \mathcal{A_{E}}(z_{t}))$. From Fig.~\ref{fig:endis}, it is clear that disentangled representation gives (1) better image quality, (2) more significant shape deformation in the younger groups, (3)  better texture transformation for older age groups, and (4) better reconfiguration. 

\begin{figure}[t!]
\centering
\vspace{-3mm}
\includegraphics[width=0.8\textwidth]{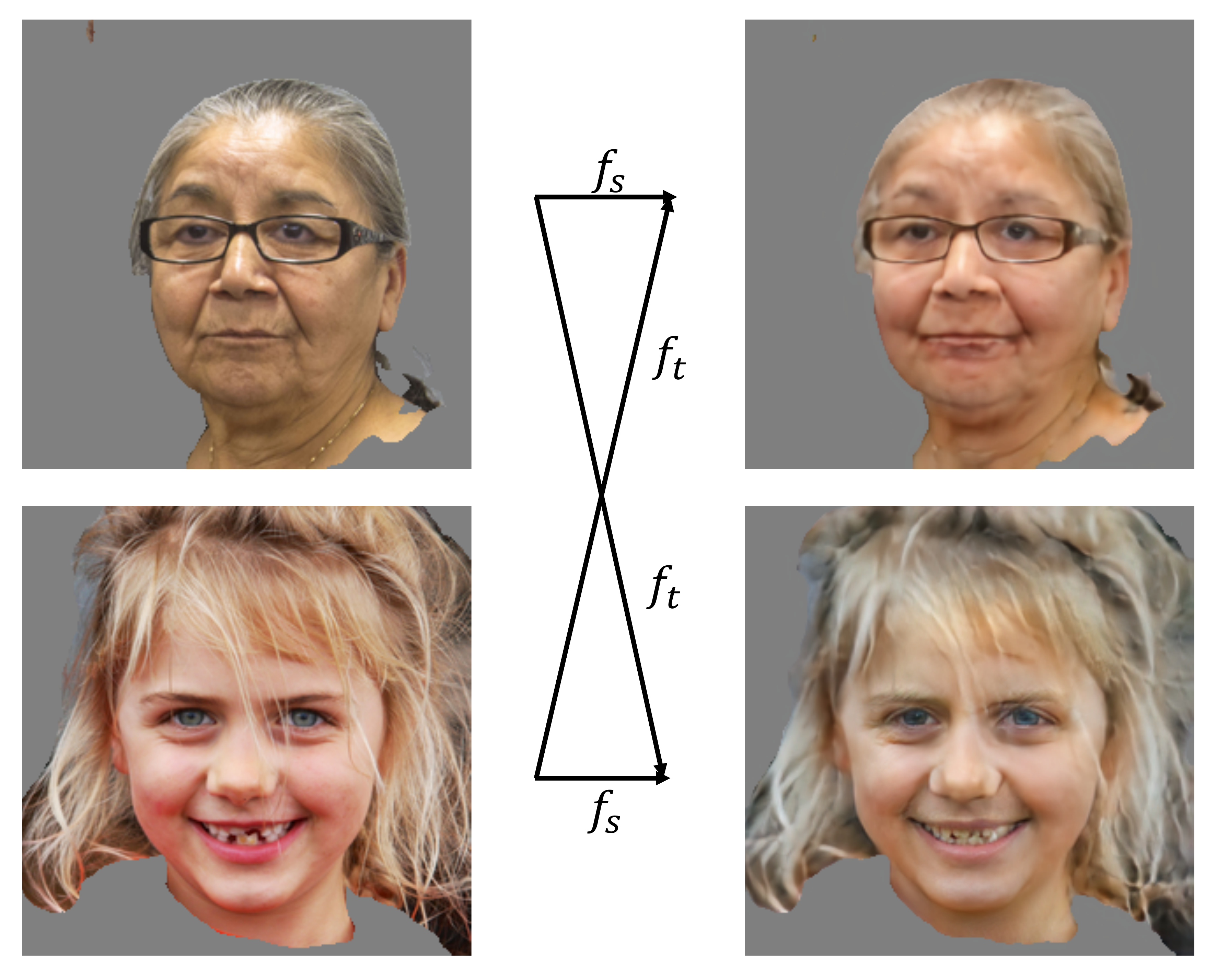}
\caption{Texture swap. In the example, the shape information ($f_{s}$) of reference image is fixed while the texture information ($f_{t}$) is swapped with another image. }
\label{fig:st_swap}
\vspace{-0.5cm}
\end{figure}

\vspace{-0.4cm}
\paragraph{The Effectiveness of Shape Regularization} To validate the effectiveness of shape regularization in Eq.~(\ref{eq:shapereg}), we trained a model without shape regularization. For comparison, we then generate the lifespan
face images by only transforming $f_{s}$. As we can see from 
Fig.~\ref{fig:wswos}, without shape regularization, there are still significant texture transformation (wrinkles) among older groups, albeit only $f_{s}$ is transformed and $f_{t}$ is fixed. On the contrary, with shape regularization,  the transformation of $f_{s}$ has nearly no effect on the texture among older groups. This suggests that our shape regularization helps to clean the texture information in $f_{s}$, thus improves disentanglement of shape and texture.
\vspace{-0.4cm}
\paragraph{Limitations} Beyond the ablation study of the disentangled representations $f_{s}$ and $f_{t}$ in controlling the aging effects as shown in Fig.~\ref{fig:shape_texture}, we further examine their limitations. A texture swap experiment is conducted, where we use one image's $f_{s}$ and the swapped $f_{t}$ from another image to generate a new image. As we can see from the result in Fig.~\ref{fig:st_swap}, $f_{t}$ seems to be dominated by skin color, which is a key defining age-agnostic texture characteristic.
As for age-related texture characteristic, e.g., wrinkles, it is not directly disentangled in $f_{t}$. However, as we can see from Fig.~\ref{fig:shape_texture}, the transformation of $f_{t}$ can amplify or suppress wrinkles. 

\vspace{-0.2cm}

\section{Conclusion}
In this paper, we proposed a novel lifespan face synthesis model based on latent representation disentanglement. In contrast to previous methods that learn entangled face representation, our method disentangles a face representation into shape and texture. We proposed age-conditioned convolution and channel attention for shape and texture transformation, respectively to reflect the distinct aging effects on shape and texture. 
Extensive experiments and evaluations show the superiority of our method compared to previous state-of-the-art models. 
\vspace{-0.2cm}
\section*{Acknowledgment}
\vspace{-0.2cm}
This work has been supported by the Federal Ministry of Education and Research (BMBF), Germany, under the project LeibnizKILabor (grant no. 01DD20003), the Center for Digital Innovations (ZDIN) and the Deutsche Forschungsgemeinschaft (DFG) under Germany’s Excellence Strategy
within the Cluster of Excellence PhoenixD (EXC 2122).

{\small
\bibliographystyle{ieee_fullname}
\bibliography{egbib}

\begin{thebibliography}{10}\itemsep=-1pt

\bibitem{antipov2017face}
Grigory Antipov, Moez Baccouche, and Jean-Luc Dugelay.
\newblock Face aging with conditional generative adversarial networks.
\newblock In {\em ICIP}, 2017.

\bibitem{arjovsky2017wasserstein}
Martin Arjovsky, Soumith Chintala, and L{\'e}on Bottou.
\newblock Wasserstein generative adversarial networks.
\newblock In {\em ICML}, 2017.

\bibitem{chen2017rethinking}
Liang-Chieh Chen, George Papandreou, Florian Schroff, and Hartwig Adam.
\newblock Rethinking atrous convolution for semantic image segmentation.
\newblock {\em arXiv preprint arXiv:1706.05587}, 2017.

\bibitem{gatys2015neural}
Leon~A Gatys, Alexander~S Ecker, and Matthias Bethge.
\newblock A neural algorithm of artistic style.
\newblock In {\em NeurIPS}, 2015.

\bibitem{goodfellow2014generative}
Ian~J Goodfellow, Jean Pouget-Abadie, Mehdi Mirza, Bing Xu, David Warde-Farley,
  Sherjil Ozair, Aaron Courville, and Yoshua Bengio.
\newblock Generative adversarial networks.
\newblock In {\em NeurIPS}, 2014.

\bibitem{he2016deep}
Kaiming He, Xiangyu Zhang, Shaoqing Ren, and Jian Sun.
\newblock Deep residual learning for image recognition.
\newblock In {\em CVPR}, 2016.

\bibitem{he2019s2gan}
Zhenliang He, Meina Kan, Shiguang Shan, and Xilin Chen.
\newblock S2gan: Share aging factors across ages and share aging trends among
  individuals.
\newblock In {\em ICCV}, 2019.

\bibitem{he2020pa}
Zhenliang He, Meina Kan, Jichao Zhang, and Shiguang Shan.
\newblock Pa-gan: Progressive attention generative adversarial network for
  facial attribute editing.
\newblock {\em arXiv preprint arXiv:2007.05892}, 2020.

\bibitem{he2019attgan}
Zhenliang He, Wangmeng Zuo, Meina Kan, Shiguang Shan, and Xilin Chen.
\newblock Attgan: Facial attribute editing by only changing what you want.
\newblock {\em TIP}, 28(11), 2019.

\bibitem{johnson2016perceptual}
Justin Johnson, Alexandre Alahi, and Li Fei-Fei.
\newblock Perceptual losses for real-time style transfer and super-resolution.
\newblock In {\em ECCV}, 2016.

\bibitem{karras2017progressive}
Tero Karras, Timo Aila, Samuli Laine, and Jaakko Lehtinen.
\newblock Progressive growing of gans for improved quality, stability, and
  variation.
\newblock In {\em ICLR}, 2018.

\bibitem{karras2019style}
Tero Karras, Samuli Laine, and Timo Aila.
\newblock A style-based generator architecture for generative adversarial
  networks.
\newblock In {\em CVPR}, 2019.

\bibitem{karras2020analyzing}
Tero Karras, Samuli Laine, Miika Aittala, Janne Hellsten, Jaakko Lehtinen, and
  Timo Aila.
\newblock Analyzing and improving the image quality of stylegan.
\newblock In {\em CVPR}, 2020.

\bibitem{kemelmacher2014illumination}
Ira Kemelmacher-Shlizerman, Supasorn Suwajanakorn, and Steven~M Seitz.
\newblock Illumination-aware age progression.
\newblock In {\em CVPR}, 2014.

\bibitem{kingma2014adam}
Diederik~P Kingma and Jimmy Ba.
\newblock Adam: A method for stochastic optimization.
\newblock In {\em ICLR}, 2015.

\bibitem{krizhevsky2012imagenet}
Alex Krizhevsky, Ilya Sutskever, and Geoffrey~E Hinton.
\newblock Imagenet classification with deep convolutional neural networks.
\newblock In {\em NeurIPS}, 2012.

\bibitem{kwak2020cafe}
Jeong-gi Kwak, David~K Han, and Hanseok Ko.
\newblock Cafe-gan: Arbitrary face attribute editing with complementary
  attention feature.
\newblock In {\em ECCV}, 2020.

\bibitem{lanitis2002toward}
Andreas Lanitis, Christopher~J. Taylor, and Timothy~F Cootes.
\newblock Toward automatic simulation of aging effects on face images.
\newblock {\em TPAMI}, 24(4), 2002.

\bibitem{lee2020maskgan}
Cheng-Han Lee, Ziwei Liu, Lingyun Wu, and Ping Luo.
\newblock Maskgan: Towards diverse and interactive facial image manipulation.
\newblock In {\em CVPR}, 2020.

\bibitem{lgfarkas}
L.G.Farkas.
\newblock {\em Anthropometry of the Head and Face}.
\newblock 2007.

\bibitem{liu2019stgan}
Ming Liu, Yukang Ding, Min Xia, Xiao Liu, Errui Ding, Wangmeng Zuo, and Shilei
  Wen.
\newblock Stgan: A unified selective transfer network for arbitrary image
  attribute editing.
\newblock In {\em CVPR}, 2019.

\bibitem{nhan2017temporal}
Chi Nhan~Duong, Kha Gia~Quach, Khoa Luu, Ngan Le, and Marios Savvides.
\newblock Temporal non-volume preserving approach to facial age-progression and
  age-invariant face recognition.
\newblock In {\em ICCV}, 2017.

\bibitem{nitzan2020face}
Yotam Nitzan, Amit Bermano, Yangyan Li, and Daniel Cohen-Or.
\newblock Face identity disentanglement via latent space mapping.
\newblock {\em TOG}, 39(6), 2020.

\bibitem{or2020lifespan}
Roy Or-El, Soumyadip Sengupta, Ohad Fried, Eli Shechtman, and Ira
  Kemelmacher-Shlizerman.
\newblock Lifespan age transformation synthesis.
\newblock In {\em ECCV}, 2020.

\bibitem{park2010age}
Unsang Park, Yiying Tong, and Anil~K Jain.
\newblock Age-invariant face recognition.
\newblock {\em TPAMI}, 32(5), 2010.

\bibitem{parkhi2015deep}
Omkar~M Parkhi, Andrea Vedaldi, and Andrew Zisserman.
\newblock Deep face recognition.
\newblock In {\em BMVC}, 2015.

\bibitem{peng2017reconstruction}
Xi Peng, Xiang Yu, Kihyuk Sohn, Dimitris~N Metaxas, and Manmohan Chandraker.
\newblock Reconstruction-based disentanglement for pose-invariant face
  recognition.
\newblock In {\em CVPR}, 2017.

\bibitem{ricanek2006morph}
Karl Ricanek and Tamirat Tesafaye.
\newblock Morph: A longitudinal image database of normal adult age-progression.
\newblock In {\em FGR06}, 2006.

\bibitem{russakovsky2015imagenet}
Olga Russakovsky, Jia Deng, Hao Su, Jonathan Krause, Sanjeev Satheesh, Sean Ma,
  Zhiheng Huang, Andrej Karpathy, Aditya Khosla, Michael Bernstein, et~al.
\newblock Imagenet large scale visual recognition challenge.
\newblock {\em IJCV}, 115(3), 2015.

\bibitem{setiawati2019bone}
Rosy Setiawati and Paulus Rahardjo.
\newblock Bone development and growth.
\newblock {\em Osteogenesis and bone regeneration}, 10, 2019.

\bibitem{shen2020interpreting}
Yujun Shen, Jinjin Gu, Xiaoou Tang, and Bolei Zhou.
\newblock Interpreting the latent space of gans for semantic face editing.
\newblock In {\em CVPR}, 2020.

\bibitem{suo2012concatenational}
Jinli Suo, Xilin Chen, Shiguang Shan, Wen Gao, and Qionghai Dai.
\newblock A concatenational graph evolution aging model.
\newblock {\em IEEE transactions on pattern analysis and machine intelligence},
  34(11):2083--2096, 2012.

\bibitem{suo2009compositional}
Jinli Suo, Song-Chun Zhu, Shiguang Shan, and Xilin Chen.
\newblock A compositional and dynamic model for face aging.
\newblock {\em TPAMI}, 32(3), 2009.

\bibitem{tazoe2012facial}
Yusuke Tazoe, Hiroaki Gohara, Akinobu Maejima, and Shigeo Morishima.
\newblock Facial aging simulator considering geometry and patch-tiled texture.
\newblock In {\em SIGGRAPH}, pages 1--1. 2012.

\bibitem{tiddeman2001prototyping}
Bernard Tiddeman, Michael Burt, and David Perrett.
\newblock Prototyping and transforming facial textures for perception research.
\newblock {\em CGA}, 21(5), 2001.

\bibitem{wang2016recurrent}
Wei Wang, Zhen Cui, Yan Yan, Jiashi Feng, Shuicheng Yan, Xiangbo Shu, and Nicu
  Sebe.
\newblock Recurrent face aging.
\newblock In {\em CVPR}, 2016.

\bibitem{wang2018face}
Zongwei Wang, Xu Tang, Weixin Luo, and Shenghua Gao.
\newblock Face aging with identity-preserved conditional generative adversarial
  networks.
\newblock In {\em CVPR}, 2018.

\bibitem{yang2018learning}
Hongyu Yang, Di Huang, Yunhong Wang, and Anil~K Jain.
\newblock Learning face age progression: A pyramid architecture of gans.
\newblock In {\em CVPR}, 2018.

\bibitem{yazici2019unusual}
Yasin Yazici, Chuan-Sheng Foo, Stefan Winkler, Kim-Hui Yap, Georgios Piliouras,
  and Vijay Chandrasekhar.
\newblock The unusual effectiveness of averaging in gan training.
\newblock In {\em ICLR}, 2019.

\bibitem{yin2019instance}
Weidong Yin, Ziwei Liu, and Chen~Change Loy.
\newblock Instance-level facial attributes transfer with geometry-aware flow.
\newblock In {\em AAAI}, 2019.

\bibitem{zhang2018unreasonable}
Richard Zhang, Phillip Isola, Alexei~A Efros, Eli Shechtman, and Oliver Wang.
\newblock The unreasonable effectiveness of deep features as a perceptual
  metric.
\newblock In {\em CVPR}, 2018.

\bibitem{zhang2017age}
Zhifei Zhang, Yang Song, and Hairong Qi.
\newblock Age progression/regression by conditional adversarial autoencoder.
\newblock In {\em CVPR}, 2017.

\bibitem{zhu2020indomain}
Jiapeng Zhu, Yujun Shen, Deli Zhao, and Bolei Zhou.
\newblock In-domain gan inversion for real image editing.
\newblock In {\em ECCV}, 2020.

\end{thebibliography}
}

\end{document}